# Hierarchical Fusion and Joint Aggregation: A Multi-Level Feature Representation Method for AIGC Image Quality Assessment

Linghe Meng, Jiarun Song
School of Telecommunications Engineering, Xidian University, Xi'an, China

*Abstract*—The quality assessment of AI-generated content (AIGC) faces multi-dimensional challenges, that span from low-level visual perception to high-level semantic understanding. Existing methods generally rely on single-level visual features, limiting their ability to capture complex distortions in AIGC images. To address this limitation, a multi-level visual representation paradigm is proposed with three stages, namely multi-level feature extraction, hierarchical fusion, and joint aggregation. Based on this paradigm, two networks are developed. Specifically, the Multi-Level Global-Local Fusion Network (MGLF-Net) is designed for the perceptual quality assessment, extracting complementary local and global features via dual CNN and Transformer visual backbones. The Multi-Level Prompt-Embedded Fusion Network (MPEF-Net) targets Text-to-Image correspondence by embedding prompt semantics into the visual feature fusion process at each feature level. The fused multi-level features are then aggregated for final evaluation. Experiments on benchmarks demonstrate outstanding performance on both tasks, validating the effectiveness of the proposed multi-level visual assessment paradigm.

## I. INTRODUCTION

Recent years have seen remarkable progress in Artificial Intelligence Generated Content (AIGC), especially in the text-to-image (T2I) field, driving the adoption of applications such as Stable Diffusion [1] and DALL-E [2], [3]. As AIGC becomes increasingly popular, effective assessment of generated image quality is essential for further model advancement.

The absence of ground-truth references in AIGC images makes Blind Image Quality Assessment (BIQA) the primary solution. Traditional BIQA methods [4]-[8] are primarily designed to detect low-level visual distortions. However, a comprehensive AIGC assessment need to consider both low-level technical quality and high-level semantic issues [9]. This dual requirement renders traditional BIQA approaches insufficient for AIGC scenarios. In response, many existing AIGCIQA methods [10]-[13] have shifted toward semantic-aware modeling, relying on deep vision models such as Vision Transformers (ViTs) to extract top-level visual features as compact vector representations of image content. While such representations are effective at capturing overall and core semantic concepts, they come at the cost of discarding multi-level visual details that are crucial for comprehensive quality assessment. Specifically, for the perceptual quality assessment, since the low-level visual information is smoothed out in the layer-by-layer abstraction, this limits the model's ability to model low-level visual defects. For the T2I correspondence, a complex text prompt typically contains information such as specific objects, visual attributes, and abstract concepts of the scene. These different levels of information may respectively correspond to different levels of feature representations in the visual encoder, and a single top-level feature may not be able to capture such rich correspondence relationships. Therefore, it is necessary to integrate visual information at multiple levels of the deep visual backbone network for comprehensive evaluation.

With this in regard, we propose an AIGC evaluation paradigm centered on multi-level visual representations. Its core process involves three stages, namely multi-level feature extraction, hierarchical feature fusion, and joint aggregation. To implement this paradigm, we propose two dedicated networks tailored for perceptual quality and T2I correspondence. For perceptual quality assessment, we propose a Multi-Level Global-Local Fusion Network (MGLF-Net). Considering that transformer-based visual backbones excel at modeling global information, while CNNs are more adept at extracting local details [14]-[18], MGLF-Net adopts a dual-backbone architecture, using both the CLIP-B/32 image encoder [19] and a ResNet50 [20], and extracts their corresponding multi-level visual features. Its core component, the Global-Local Fusion Block(GLF Block), uses learnable queries to efficiently fuse the complementary global and local information from both backbones at each visual level. For T2I correspondence task, we design a Multi-Level Prompt-Embedded Fusion Network (MPEF-Net). Pre-trained on large-scale image-text pairs, CLIP has constructed a unified visual-language space, which makes it well-suited for this task. Therefore, MPEF-Net selects CLIP as its backbone. Its core component, the Prompt-Embedded Fusion Block(PEF Block), embeds the high-level prompt semantic into the hierarchical visual fusion process, helping the learnable queries to precisely query and fuse prompt-relevant visual information at different visual levels. After hierarchical fusion, both networks aggregate the multi-level features into a holistic representation for final prediction. The main contributions of this paper are as follows:

- We propose an AIGCIQA evaluation paradigm centered on multi-level visual representations. This paradigm deconstructs the evaluation process into three stages—multi-level feature extraction, hierarchical fusion, and joint aggregation. By integrating a complete hierarchy of visual information from low to high levels, the proposed paradigm overcomes the limitations of relying on a single top-level feature and thereby enables a more comprehensive assessment of AIGC.

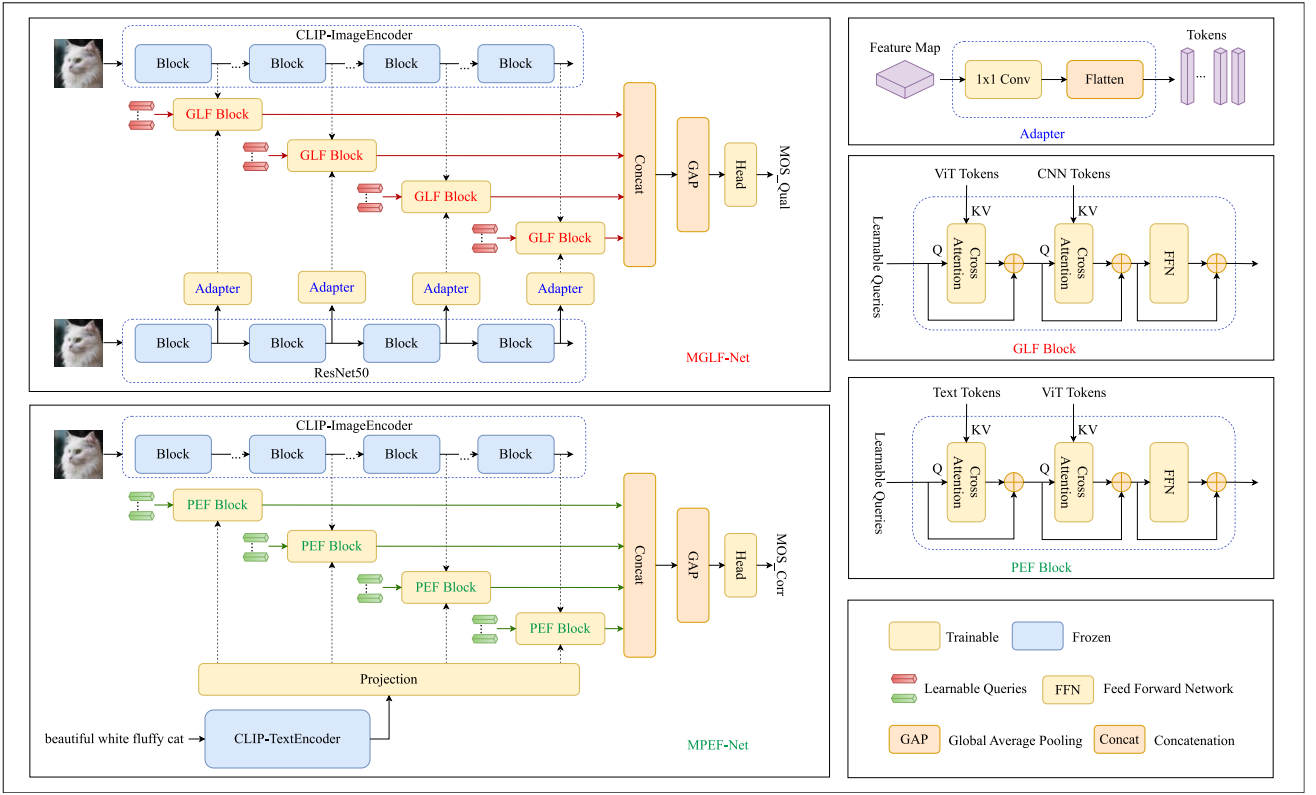

Fig. 1. An overview of our proposed framework and its instantiation networks for two tasks: MGLF-Net (upper left) hierarchically fuses global features and local features; MPEF-Net (lower left) utilizes Prompt embedding to assist in modeling image-text consistency at multiple visual representation levels.

- We design task-specific multi-level fusion networks for perceptual quality and T2I correspondence. The MGLF-Net effectively fuses global and local information across different visual levels. The MPEF-Net uses a prompt-embedding fusion mechanism to model T2I semantic correspondence across multiple representation levels. Experiments on benchmark datasets show that proposed methods achieve outstanding performance in both tasks, validating effectiveness of proposed paradigm.

## II. METHODOLOGY

Framework of the proposed method is illustrated in Fig. 1, which is centered on multi-level visual representations. The framework is designed to comprehensively assess AIGC image quality by processing and integrating features extracted from multiple levels of the visual backbone.

### A. Multi-Level Feature Extraction

Diverging from existing methods that rely on a single, top-level feature from the backbone, our work is fundamentally motivated by the need to integrate a rich hierarchy of features, enabling a holistic judgment across multiple levels. The architecture takes the AIGC image *Image* and the corresponding text prompt *Prompt* as inputs. To ensure hierarchical diversity in our features, we have devised the following extraction strategy:

**Transformer feature** For the CLIP image encoder (which contains 12 Transformer Blocks), we select the outputs of its 3rd, 6th, 9th, and 12th layers as visual representations for four distinct levels of information abstraction, which aims to balance feature diversity with representational power. It not only incorporates the strong global representations from the final layer, but also integrates features from multiple intermediate layers, leading to richer and more comprehensive information. The outputs of the image encoder used for image representation are denoted as $G_i \in \mathbb{R}^{B \times N_I \times D}$. For the CLIP text encoder, we only require the token sequence from its final layer as the semantic representation for the *Prompt*. Since the internal embedding dimensions of the image encoder and text encoder are inconsistent, the text token sequence is first processed by a unified projection before being fed into the PEF Block, the projected features are denoted as $P \in \mathbb{R}^{B \times N_P \times D}$. Here, $N_I$ and $N_p$ are respectively the number of tokens corresponding to each level of the image encoder and the text encoder, $D$ is the internal embedding dimension of the Image Encoder, and $B$ is the batch size.

**CNN feature** In MGLF-Net, we extract feature maps from stages 1 through 4 of a ResNet50. Since the CNN's feature maps are 2D, they must be transformed to enable interaction with the

1D token sequence from the CLIP image encoder. This is achieved using an Adapter module(as shown on the right side of Fig. 1), which converts them into a 1D feature sequence. After the transformation, the different levels of 1D feature sequences are denoted as $L_i$, which is defined as:

$$L_i = Adapter\left(ResNet_{satge_i}\left(Image\right)\right) \in R^{B \times C_i \times D} \quad (1)$$

where $C_i$ represents the number of tokens corresponding to each CNN level after passing through the Adapter. The index $i \in [0, 1, 2, 3]$ represents the four corresponding visual levels for both the Transformer and CNN features.

### B. Hierarchical Fusion

The features at different levels of deep visual networks exhibit complexity and representational differences. Directly passing them to the final aggregation and prediction stage may introduce redundancy and complicate the decision-making process. To address this, we refine features at each level during the hierarchical fusion stage, enabling more effective downstream aggregation. Specifically, we adopt a learnable query-based strategy, which is trained end-to-end to adaptively select and fuse the most relevant information at each level for the current evaluation task.

Both our GLF Block in MGLF-Net and Prompt-Embedded Fusion PEF Block in MPEF-Net adopt a unified multi-stage query refinement structure. At each level $i \in [0, 1, 2, 3]$, a set of learnable queries $Q_i \in \mathbb{R}^{B \times N_Q \times D}$, where $N_Q$ is the configurable number of learnable queries, is progressively refined through (1) a first-stage cross-attention to integrate global information (or prompt semantics), (2) a second-stage cross-attention to attend to relevant local details (or prompt-relevant visual information), and (3) a Feed-Forward Network (FFN) acting as a non-linear transformation to produce the final representation $\tilde{Q}_i$ at each level $i$. Each stage includes residual connections to ensure stable information transmission. This general process can be formulated as:

$$Q_i' = CrossAttention(Q_i, K_i, V_i) + Q_i \quad (2)$$

$$Q_i'' = CrossAttention(Q_i', K_i', V_i') + Q_i' \quad (3)$$

$$\tilde{Q}_i = FFN(Q_i'') + Q_i'' \quad (4)$$

where $(K_i, V_i)$ and $(K_i', V_i')$ denote different conditioning inputs depending on the task. Specifically, in GLF Block, $(K_i, V_i)$ **denotes** $(G_i, L_i)$, $Q_i$ first absorb the global information by performing cross-attention with the global features $G_i$, then these globally-aware queries are further enhanced by attending to the local details from the features $L_i$, in PEF Block, $(K_i', V_i')$ **denotes** $(P, G_i)$, $Q_i$ first interact with the prompt's semantic representation $P$ to absorb the prompt semantic, then these queries perform cross-attention with visual information $G_i$ to extract prompt-relevant visual information.

### C. Joint Aggregation and Regression

Following the steps above, a set of task-enhanced queries $\tilde{Q}_i$, each $\tilde{Q}_i$ is a condensed representation of the information from its respective level. These queries are first concatenated along the token dimension and then processed by a Global Average Pooling (GAP) operation to aggregate the token sequence into a single, fixed-dimensional feature vector $F_{task}$.

$$\tilde{Q}_{cat} = Concat\left(\tilde{Q}_0, \ldots, \tilde{Q}_3\right) \in \mathbb{R}^{B \times 4N_Q \times D} \quad (5)$$

$$F_{task} = GAP(\tilde{Q}_{cat}) \in R^{B \times D} \quad (6)$$

Finally, $F_{task}$ is fed into an MLP regression head to predict the final quality score, MOS (Mean Opinion Score), for the corresponding task.

$$MOS = MLP(F_{task}) \quad (7)$$

### III. EXPERIMENTS

This section will present the datasets, evaluation metrics, implementation details, and experimental results.

**Datasets** We validate our work on three widely-used AIGCIQA datasets: AGIQA-1K [21], AGIQA-3K [22], and AIGCIQA2023 [23]. The AGIQA-1K dataset contains 1,080 images generated by two text-to-image (T2I) models, each annotated with a Mean Opinion Score (MOS). AGIQA-3K expands this to 2,982 images from six T2I models, offering MOS labels for both perceptual quality and T2I correspondence. The AIGCIQA2023 dataset comprises 2,400 images produced by six T2I models, with MOS labels provided for perceptual quality, authenticity, and correspondence dimensions.

**Evaluation Metrics** We use Spearman's Rank Correlation Coefficient (SRCC) and the Pearson Linear Correlation Coefficient (PLCC), which are standard metrics for quality assessment. They respectively measure the ranking capability and fitting accuracy of the model's predictions.

**Implementation Details** All experiments were conducted using Python 3.8, PyTorch 2.0.0, and CUDA 11.8 on a single NVIDIA GeForce RTX 4090 GPU. CLIP-B/32 and ResNet50 were adopted as backbone networks. Multi-level features were extracted in parallel from the 3rd, 6th, 9th, and 12th layers of the CLIP image encoder. Models were trained for 30 epochs with a batch size of 16 using the AdamW optimizer (learning rate: 1e-5, weight decay: 1e-5).

### A. Main Results

To demonstrate the effectiveness of our approach, we compare it against several traditional IQA methods [24]-[28] and the latest AIGC-IQA methods [10]-[12] [29]-[32]on the AGIQA-1K, AGIQA-3K, and AIGCIQA2023 datasets. Each dataset was split into training and test sets using an 8/2 ratio. We maintained a fixed random seed throughout the data splitting and model training processes to ensure reproducibility.

TABLE I - TABLE III respectively present the experimental results on AIGCIQA2023, AGIQA-3K and AGIQA-1K, where Qual. and Corr. respectively represent perceptual quality and

TABLE I
THE COMPARISON RESULTS ON AIGCIQA2023 DATASET. **BOLD** AND <u>UNDERLINED</u> VALUES RESPECTIVELY INDICATE THE BEST AND SECOND-BEST RESULTS.

| Method | Qual. | | Auth. | | Corr. | |
|---|---|---|---|---|---|---|
| | SRCC | PLCC | SRCC | PLCC | SRCC | PLCC |
| DBCNN [24] | 0.8339 | 0.8577 | 0.7485 | 0.7436 | 0.6837 | 0.6787 |
| StairIQA [27] | 0.8264 | 0.8483 | 0.7596 | 0.7514 | 0.7176 | 0.7133 |
| HyperIQA [26] | 0.8357 | 0.8504 | 0.7758 | 0.7790 | 0.7318 | 0.7222 |
| PSCR [32] | 0.8371 | <u>0.8588</u> | 0.7828 | 0.7750 | 0.7465 | 0.7397 |
| CLIP-AGIQA [11] | 0.8140 | 0.8302 | <u>0.7940</u> | <u>0.7797</u> | - | - |
| AMFF-Net [12] | <u>0.8409</u> | 0.8537 | 0.7782 | 0.7638 | <u>0.7749</u> | <u>0.7643</u> |
| Our | **0.8499** | **0.8664** | **0.7992** | **0.7922** | **0.7764** | **0.7649** |

TABLE II
THE COMPARISON RESULTS ON AGIQA-3K DATASET. **BOLD** AND <u>UNDERLINED</u> VALUES RESPECTIVELY INDICATE THE BEST AND SECOND-BEST RESULTS.

| Method | Qual. | | Corr. | |
|---|---|---|---|---|
| | SRCC | PLCC | SRCC | PLCC |
| DBCNN [24] | 0.8154 | 0.8747 | 0.6329 | 0.7823 |
| MGQA [25] | 0.8283 | 0.8944 | 0.7244 | 0.8430 |
| HyperIQA [26] | 0.8526 | 0.8975 | 0.7437 | 0.8471 |
| IP-IQA [29] | 0.8634 | 0.9116 | 0.7578 | 0.8544 |
| IPCE [10] | 0.8841 | 0.9266 | 0.7697 | 0.8825 |
| MoE-AGIQA-v2 [30] | 0.8746 | 0.9282 | - | - |
| SF-IQA [31] | <u>0.9024</u> | **0.9314** | **0.8454** | **0.9072** |
| Our | **0.9039** | <u>0.9310</u> | <u>0.8410</u> | <u>0.8968</u> |

TABLE III
THE COMPARISON RESULTS ON AGIQA-1K DATASET. **BOLD** AND <u>UNDERLINED</u> VALUES RESPECTIVELY INDICATE THE BEST AND SECOND-BEST RESULTS.

| Method | Qual. | |
|---|---|---|
| | SRCC | PLCC |
| DBCNN [24] | 0.7491 | 0.8211 |
| HyperIQA [26] | 0.7803 | 0.8299 |
| CONTRIQUE [28] | 0.8073 | 0.8866 |
| PSCR [32] | 0.8430 | 0.8403 |
| IP-IQA [29] | 0.8401 | **0.8922** |
| IPCE [10] | <u>0.8535</u> | 0.8792 |
| MoE-AGIQA-v2 [30] | 0.8501 | **0.8922** |
| Our | **0.8648** | <u>0.8874</u> |

TABLE IV
ABALATION STUDIES ON THE AGIQA-3K DATASET. THE BEST RESULTS ARE **BOLDED**

| Model settings | Qual. | | Corr. | |
|---|---|---|---|---|
| | SRCC | PLCC | SRCC | PLCC |
| w/o Transformer features | 0.8615 | 0.9043 | - | - |
| w/o CNN features | 0.8945 | 0.9241 | - | - |
| w/o Prompt-Embedded | - | - | 0.7849 | 0.8710 |
| Single-Level (the last) | 0.8902 | 0.9211 | 0.8258 | 0.8869 |
| Full Model | **0.9039** | **0.9310** | **0.8410** | **0.8968** |
| 4 queries | **0.9039** | **0.9310** | 0.8363 | 0.8902 |
| 8 queries | 0.8975 | 0.9275 | **0.8410** | **0.8968** |

T2I correspondence. The comparative results clearly show the effectiveness of our proposed framework.

*B. Ablation Study*

To evaluate the contribution of each key component in our proposed methods, we conducted a comprehensive ablation study on the AGIQA-3K dataset for both the perceptual quality and text-image consistency assessment tasks. The results are summarized in Table IV.

Specifically, **w/o Transformer features** refers to removing the Transformer features from the GLF Block, relying solely on CNN features. Conversely, **w/o CNN features** removes the CNN features, using only the Transformer features. The results for these variants indicate that both the global information from the Transformer architecture and the fine-grained local details from the CNN architecture are crucial for MGLF-Net. **w/o Prompt-Embedded** denotes the removal of the prompt's semantic features in the PEF Block. The performance drop confirms the effectiveness of using these semantics to help the Learnable Queries in extracting consistency-relevant image features. Notably, **Single-Level (the last)** represents processing only the features from the final visual level (i.e., the top-level feature) for both tasks. The results underscore the importance of our multi-level approach, demonstrating that multi-level feature extraction and hierarchical fusion yield a more comprehensive information representation beneficial to both evaluation tasks. Finally, the results for **4 queries** and **8 queries** compare the performance with different numbers of learnable queries in the fusion blocks. The study shows that using 4 queries is optimal for the GLF Block, whereas 8 queries yield better performance for the PEF Block.

## IV. CONCLUSION

In this paper, we propose a unified AIGCIQA paradigm based on multi-level visual representations to overcome the limitations of the approaches that rely on a single, top-level visual feature. It consists of three stages: multi-level feature extraction, hierarchical fusion, and joint aggregation. We design MGLF-Net for perceptual quality assessment by fusing global-local features across layers, and MPEF-Net for T2I alignment by embedding prompt semantics into each visual fusion stage. The aggregated multi-level features form a more comprehensive representation for final prediction. Experiments on multiple benchmarks confirm the effectiveness of our approach across both tasks.